\documentclass[letterpaper, 10 pt, conference]{ieeeconf}  

\usepackage{url}
\usepackage{graphicx}
\usepackage{booktabs}
\usepackage{romannum}
\usepackage{amsfonts}
\usepackage{multirow}
\usepackage[table,xcdraw]{xcolor}
\usepackage{array}
\usepackage{float}
\usepackage[export]{adjustbox} 
\usepackage{algorithm}
\usepackage{balance}
\usepackage{cite}
\usepackage{algpseudocode} 
\usepackage{longtable}
\usepackage{booktabs}

\usepackage{graphics}
\usepackage{upgreek}
\usepackage{amssymb}
\usepackage{refstyle}

\usepackage{amsmath}
\usepackage{subfigure}
\usepackage{todonotes}
\usepackage{mathtools}

\usepackage{hyperref}

\usepackage{cleveref}

\newcommand\copyrighttext{%
  \footnotesize \textcopyright 2023 IEEE. Personal use of this material is permitted.
  Permission from IEEE must be obtained for all other uses, in any current or future
  media, including reprinting/republishing this material for advertising or promotional
  purposes, creating new collective works, for resale or redistribution to servers or
  lists, or reuse of any copyrighted component of this work in other works.
}
\newcommand\copyrightnotice{%
\begin{tikzpicture}[remember picture,overlay]
\node[anchor=south,yshift=10pt] at (current page.south) {\fbox{\parbox{\dimexpr\textwidth-\fboxsep-\fboxrule\relax}{\copyrighttext}}};
\end{tikzpicture}%
}

\IEEEoverridecommandlockouts                              

\title{\LARGE \bf
A Collision-Aware Cable Grasping Method \\in Cluttered Environment}
\author{Lei Zhang$^{1,2}$, Kaixin Bai$^{1,2}$, Qiang Li$^{1*}$, Zhaopeng Chen$^{2*}$, Jianwei Zhang$^{1}$
\thanks{*Corresponding authors.}
\thanks{{$^{1}$TAMS (Technical Aspects of Multimodal Systems), Department of
Informatics, Universit\"at Hamburg}. {$^{2}$Agile Robots AG, Germany}.}
\thanks{** This research has received funding from the German Research Foundation (DFG) and the National Science Foundation of China (NSFC) in project Crossmodal Learning, DFG TRR-169/NSFC 61621136008, DEXMAN project (410916101), partially supported by European projects H2020 STEP2DYNA (691154) and ULTRACEPT (778602).}
}
\DeclarePairedDelimiterX{\norm}[1]{\lVert}{\rVert}{#1}

\begin{document}

\maketitle
\copyrightnotice

\thispagestyle{empty}
\pagestyle{empty}

\begin{abstract}    

We introduce a Cable Grasping-Convolutional Neural Network (CG-CNN) designed to facilitate robust cable grasping in cluttered environments. Utilizing physics simulations, we generate an extensive dataset that mimics the intricacies of cable grasping, factoring in potential collisions between cables and robotic grippers. We employ the Approximate Convex Decomposition technique to dissect the non-convex cable model, with grasp quality autonomously labeled based on simulated grasping attempts. The CG-CNN is refined using this simulated dataset and enhanced through domain randomization techniques. Subsequently, the trained model predicts grasp quality, guiding the optimal grasp pose to the robot's controller for execution. Grasping efficacy is assessed across both synthetic and real-world settings. Given our model's implicit collision sensitivity, we achieved commendable success rates of 92.3\% for known cables and 88.4\% for unknown cables, surpassing contemporary state-of-the-art approaches. Supplementary materials can be found at \href{https://leizhang-public.github.io/cg-cnn/}{https://leizhang-public.github.io/cg-cnn/}.
\end{abstract}

\section{Introduction}
\label{intro}
In industrial settings, humans primarily perform cable grasping. The inherent challenge lies in the fact that cables are deformable, non-standard components showcasing intricate geometric characteristics. Their pliability, combined with complex geometry, makes the automation of grasping these components a notable challenge in robotic applications. When introducing a scenario where a robot is required to grasp stacked cables, the complexity escalates. Robots must meticulously address the mutual occlusion stemming from cables in varied geometric positions and forms. Advanced robotic manipulation, fortified with collision awareness, becomes paramount to prevent potential damage to products within containers, safeguarding their economic value.

\begin{figure}[htbp]
	\begin{center}
  		\includegraphics[width=8cm]{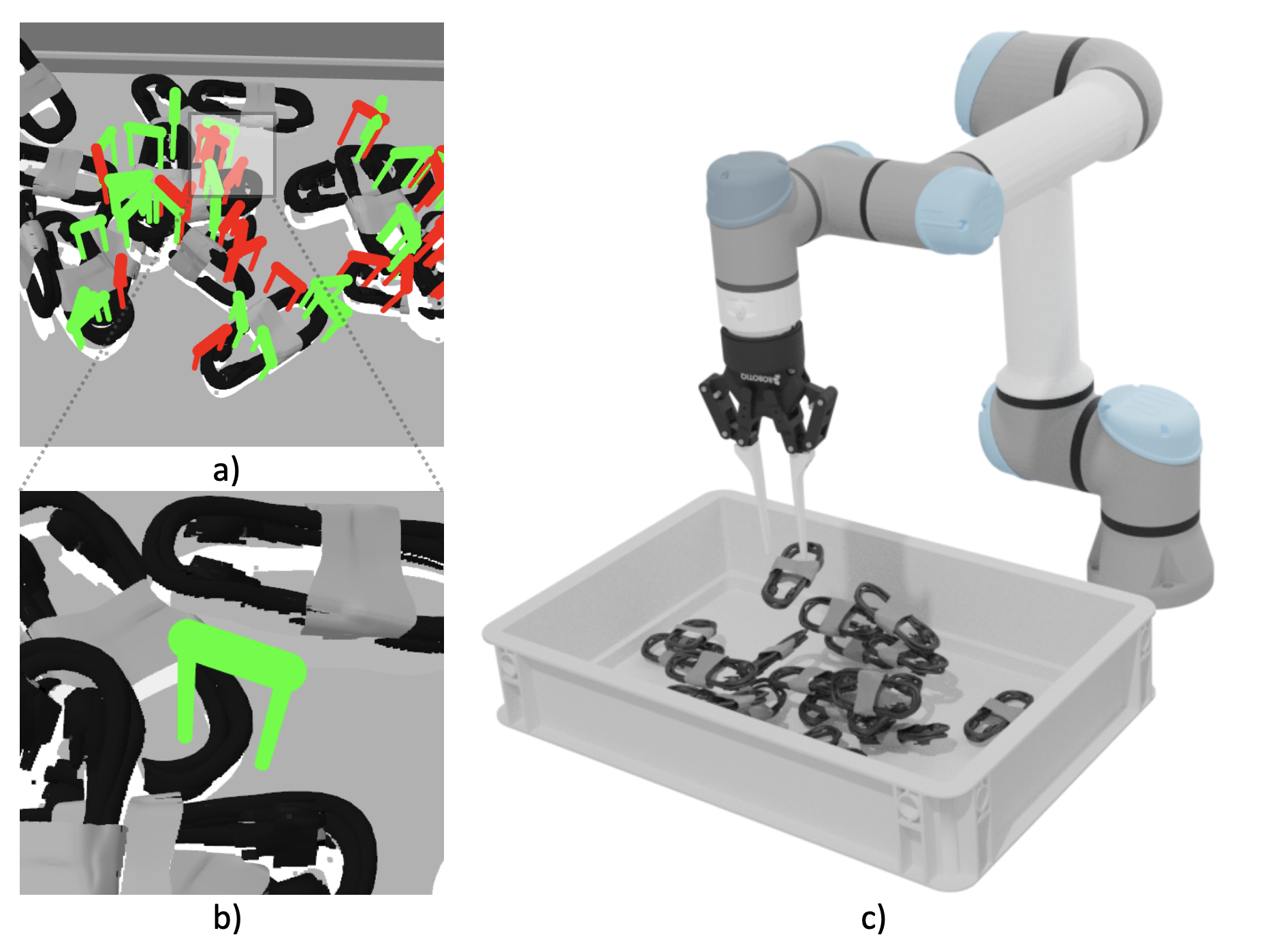}
    		\caption{Grasping cables from cluttered scenes exploiting the grasping samples approach. The grasp candidates are obtained considering the force closure principle. a) CG-CNN estimates grasp quality of grasp candidates. Positive grasp candidates are in green: quality $> 0.5$; Negative grasp candidates are in red : quality $< 0.5$. b) Optimal grasp candidate. c) Physics simulation for grasping. }
		\label{fig.factory_scene_and_simtoreal_scene}
	\end{center}
\end{figure}
\setlength{\textfloatsep}{5pt}

Promising strides in cable grasping have been observed using model-free and learning-based strategies, particularly evident in their successful applications for grasping rigid objects. Pioneering with the traditional force closure technique \cite{nguyen1987constructing} and progressing with the Ferrari-Canny metric \cite{ferrari1992planning}, model-free grasping methods have been pivotal in generating optimized grasp poses, leveraging vast data reservoirs \cite{liang2019pointnetgpd,mahler2019learning,fang2020graspnet,morrison2020egad}. Such datasets can either be architected via analytical techniques \cite{mahler2017dex} or honed through trial-and-error in simulations \cite{zhao2022learning,cai2019metagrasp,kiatos2022learning,zeng2018learning,gilles2022metagraspnet,eppner2021acronym}. Capitalizing on these datasets, several networks have emerged, aiming to plan optimal grasps for diverse objects and settings \cite{mahler2017dex,fang2020graspnet,sundermeyer2021contact}. Nevertheless, the task becomes daunting when dealing with flat, non-convex entities like cables, especially when ensuring they remain collision-free with the ambient environment. While previous methods have achieved high grasping success rates on rigid convex objects, their performance considerably diminishes when it comes to cable grasping. This decrement in efficacy can be attributed to two primary reasons: Firstly, the intricate geometric features of cables are notably distinct from those of objects present in standard datasets. Secondly, the intertwined nature of cables presents a unique challenge; algorithms that fail to account for potential collision relationships often result in grasping failures due to entanglements and collisions. This predicament led to the genesis of our proposed CG-CNN.

Embracing model-free techniques for cable grasping in cluttered environment (e.g. Fig.~\ref{fig.factory_scene_and_simtoreal_scene}) necessitates grappling with challenges such as the cable's dynamic geometric attributes, stacking nuances, robotic grasp efficiency, and collision cognizance. Our method integrates a grasp sampling algorithm anchored on the force closure principle, capturing grasp candidates from depth images. The training phase of CG-CNN encapsulates implicit collision awareness. Here, the intricate act of grasping the non-convex cable is mirrored in a physics simulation, underpinned by domain randomization techniques, accommodating potential collisions with both the encompassing environment and other cables in proximity. The simulation discerns the viability of each grasp—whether it's positive or negative. CG-CNN then derives this intricate relationship, mapping directly from depth imagery (focused on the grasp candidate vicinity) to the grasp quality. The crux of our paper can be distilled into three core contributions:

The core contributions of this paper are summarized as follows.
\begin{enumerate}
\item Introduction of a pioneering, deep learning-fueled cable grasping method incorporating collision-sensitive CG-CNN, adept at forecasting grasp qualities while inherently recognizing and navigating collision scenarios, enhancing the grasping phase's safety quotient.
\item Crafting expansive datasets tailored for cable grasping, enhancing the breadth and depth of contemporary model-free robotic grasping paradigms.

\item Our cable grasping method outperforms SOTA (state-of-the-art) methods in cable grasping from clutter and delivers performance - success benchmark of 92.3\% and 88.4\% respectively in known and unknown cable grasping from clutter.  
\end{enumerate}

\section{Related Work}
\label{sec:relatedwork}
\subsection{Cable Grasping}

Cable assembling is a challenge robotic manipulation task. Many significant progresses have been achieved in cable manipulation~\cite{caporali2021combining}, cable insertion~\cite{she2021cable}, and rearrangements~\cite{seita2021learning}. Deformable object manipulation has been partially investigated by predicting robot actions through the utilization of visual data~\cite{seita2020deep,seita2019deep}. Cable grasping with complex geometries in cluttered environment, however, remains relatively unexplored~\cite{zhang2022learning}, which will be the focus of this paper. Poor visual prediction still limits performance of cable grasping in heavily occluded clutter~\cite{zhang2022learning}. Motivated by this, we learn and generalize robotic policy with collision awareness in simulation using domain randomization. 

\subsection{Model-free Robotic Grasping}

The robotic grasping can be classified into three categories: model-based, half-model-based, model-free robotic grasping. Model-based robotic grasping estimates the pose of the object and performs grasping using sampling algorithms or predefined grasping poses~\cite{fu20226d,konishi2019real}. Half model-based grasping methods usually extract the regions of interest with similar features then execute grasping~\cite{lin2022primitive}. Model-based grasping methods necessitate the utilization of 3D models for each object under consideration, whereas model-free methods circumvent this requirement by directly inferring the grasping poses with the trained models. Analytical grasp sampling methods and Ferrari-Canny metrics are popular approaches for model-free grasping from cluttered scenes and corresponding dataset generation \cite{mahler2017dex,mahler2019learning,liang2019pointnetgpd,fang2020graspnet,sundermeyer2021contact}. Affordance-based grasping frameworks \cite{zeng2018robotic,kumra2022gr,morrison2018closing,zhang2022towards} also represent a type of model-free grasping.

In robotic grasping domain, some representative model-free approaches are Dex-Net~\cite{mahler2017dex,mahler2019learning}, GraspNet-1Billion~\cite{fang2020graspnet} and Contact-GraspNet~\cite{sundermeyer2021contact}. Dex-Net 2.0 primarily generated grasp candidates from scene featuring single objects rather than cluttered scenarios. It trained a Grasp Quality Convolutional Neural Network (GQ-CNN) for grasping in cluttered environments~\cite{mahler2017dex,mahler2019learning}. Similarly, GraspNet-1Billion~\cite{fang2020graspnet} sampled grasp candidates in single-object scenes and applied data augmentation to simulate cluttered scenes, subsequently proposed a large-scale benchmark for grasping general objects. However, neither Dex-Net nor GraspNet-1Billion incorporated collision simulations between the end effector and the surrounding environment during grasp generation and evaluation. As a result, the collision information within the grasp samples produced by these methods is not considered reliable. Contact-GraspNet~\cite{sundermeyer2021contact} also leveraged analytical computation to consider collision in the generation of grasp samples within their datasets and trained grasp networks that possess implicit collision awareness. However, Contact-GraspNet is not able to avoid grasping multiple cables from cluttered environments because it recognizes neighboring cables as different parts of the same object. To solve these problems, we propose CG-CNN to grasp single cable from complicated scenes without multi-cable grasping and dropping after grasping.

We have identified three aforementioned methods, including Dex-Net~\cite{mahler2017dex,mahler2019learning}, GraspNet-1Billion~\cite{fang2020graspnet}, and Contact-GraspNet~\cite{sundermeyer2021contact}, that rely on analytical approaches for dataset generation. However, they lack sample validation within a virtual environment. Additionally, analytical methods entail a high computational cost. Recent studies have also substantiated the reliability and high efficiency of trial-and-error simulations of complicated environments~\cite{zhao2022learning,cai2019metagrasp,kiatos2022learning,zeng2018learning,gilles2022metagraspnet,eppner2021acronym}. Although simulations are deemed more reliable, none of the three methods has evaluated the simulation results of grasping samples in cluttered scenarios. To enhance the reliability of grasping samples within current dataset methodologies, we introduce a novel simulation environment for simulating grasping single cable in cluttered settings. We validate grasping samples in cluttered scenes through simulation, leading to more dependable labels compared to previous approaches. Furthermore, the grasping outcomes can be automatically annotated through the trial-and-error approach. In summary, our CG-CNN aims to more reliably and efficiently advance the assessment of grasping in cluttered environments, thereby enabling the application of this method in real industrial settings.

\section{Problem Statement and Methods}\label{sec:method}

\subsection{Problem Statement}

In cluttered environment, grasping cables with complicated shapes requires collision awareness. The robot needs to find graspable positions in an unobstructed area and grasp only one cable from stacked cables.

\begin{figure}[htbp]
	\begin{center}
		\includegraphics[width=5.5cm]{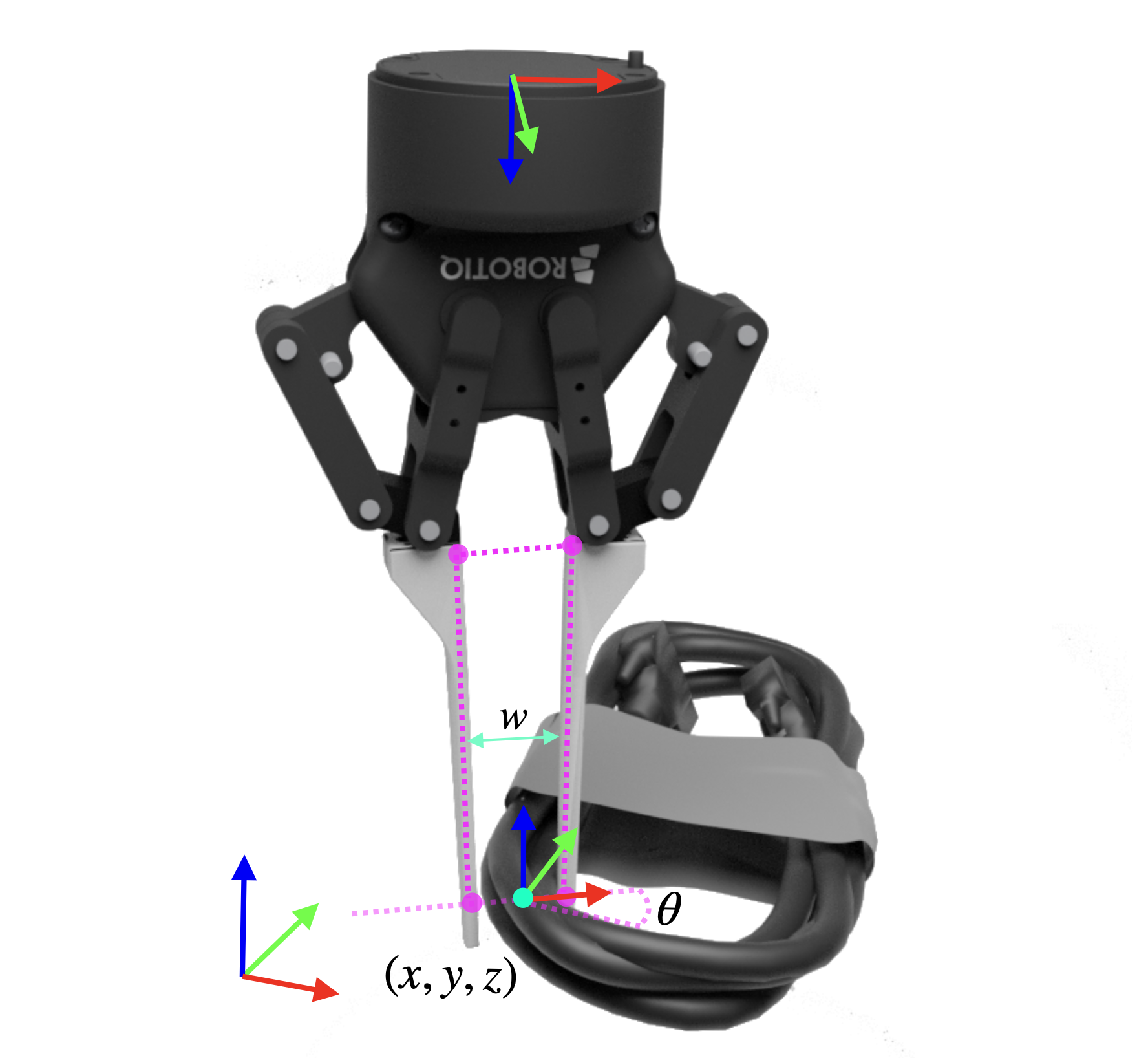}
		\caption{Definition of the grasp pose.}
		\label{fig.grasp_configuration}
	\end{center}
\end{figure}
\begin{figure*}[htbp]
	\begin{center}
  		\includegraphics[width=17.5cm]{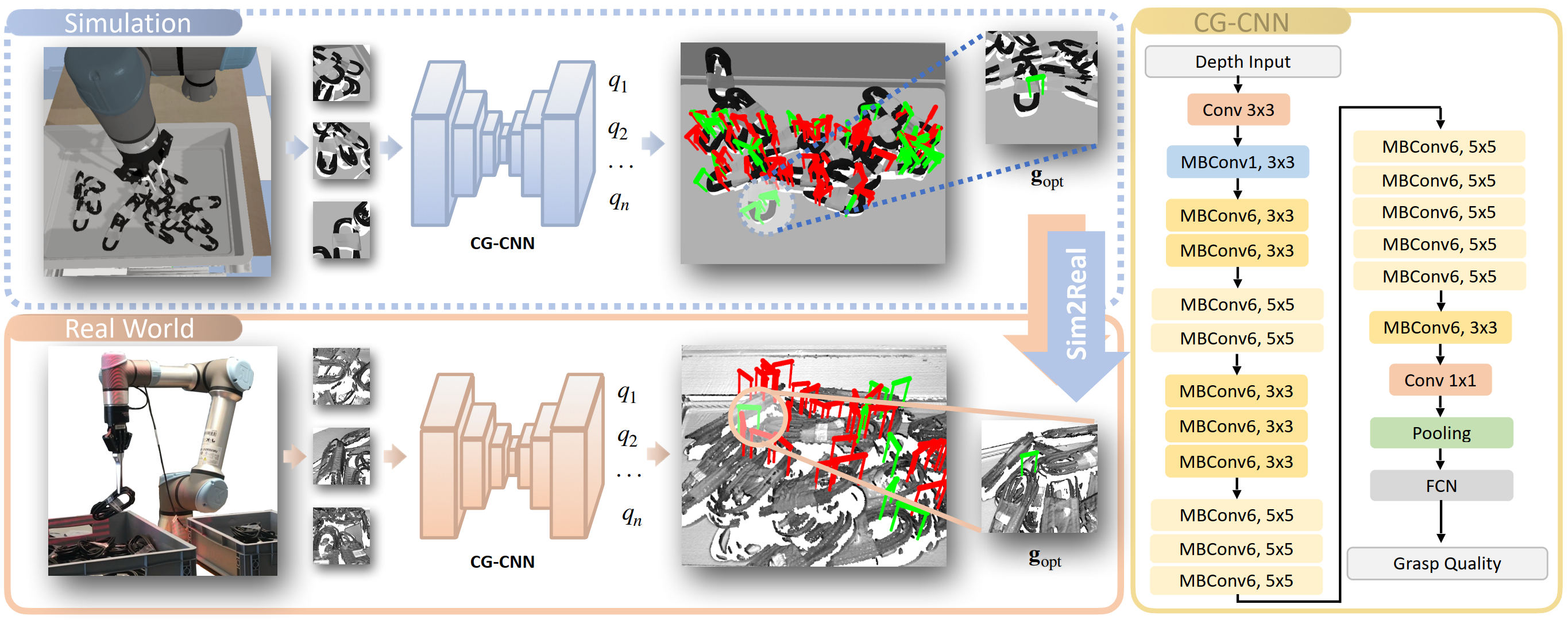}
  \caption{CG-CNN based grasping pipeline (left) and CG-CNN network architecture (right). The grasp sampling method is used to sample grasp candidates of two-jaw gripper from depth image. CG-CNN learns grasp policy $\pi_{\rm CG-CNN}$ by predicting grasp qualities using the cropped depth images of sampled grasp candidates. The optimal grasping pose $g_{\rm{opt}}$ is selected and implemented in simulation and real world based on grasp strategy.}
		\label{fig.pipeline_proposed_method}
	\end{center}
\end{figure*}

A two-jaw gripper is used for grasping the cables in a horizontal pose from a deep box, as shown in Fig.~\ref{fig.grasp_configuration}. The grasp pose is 
\begin{equation}
g = [x,y,z,\theta,w]
\end{equation}

which includes grasp position $(x, y, z)$, grasp orientation $\theta$, and grasp width $w$.

In order to figure out the optimal grasp pose in the cluttered environment, we start with simulating the non-convex cable in our proposed simulator. Convex decomposition approach is utilized to process the geometric models, which will be introduced in Sec.~\ref{sec.convex_decomposition}. 
We employ force closure grasping sampling method to find N valid grasp candidates $A=[\textbf{g}_{0}, \textbf{g}_{1},\ldots,\textbf{g}_{\rm N-1}]$ based on the depth image, as detailed in Sec.~\ref{sec.force_closure}. The executed grasp width $w_{\rm estimated}$ is estimated analytically according to two contact points in the grasp sampling.

The collision-aware cable grasping simulation is used for data collection, and the collected data is employed in training the CG-CNN for estimating the grasp quality (probability of successful grasping). The whole pipeline is shown in Fig.~\ref{fig.pipeline_proposed_method} and detailed in Sec.~\ref{sec.learn_cnn_from_simulation}. In order to improve the computational efficiency, we only use the cropped depth image $\boldsymbol I^{\rm c}$ of the grasp candidate $\textbf{g}_{i}$ for training the network. Using the trained network, the grasp quality can be inferred given a new stacked cable image. Based on the inference, the optimized grasp posture is computed from Eq.~\ref{equ:optgrasp}.

\begin{equation}
\label{equ:optgrasp}
\begin{split}
&\bigcup_{i=0}^{N-1} q_{i} = \bigcup_{i=0}^{N-1} f_{\rm CG-CNN} (\boldsymbol I^{c}_{i})\\
&\textbf{g}_{\rm opt}  = h_{\rm grasp-strategy} (\bigcup_{i=0}^{N-1} q_{i})\\
\end{split}
\end{equation}

where $\bigcup \limits_{i=0}^{N-1} \textbf{I}^{c}_{i}$ indicates cropped depth images of grasp candidates, $\bigcup \limits_{i=0}^{N-1} q_{i}$ denotes inferred qualities based on CG-CNN $f_{\rm CG-CNN}$, and $h_{\rm grasp-strategy}$ represents grasp strategy for selecting optimal grasp candidate $\textbf{g}_{\rm opt}$, as described in Sec.~\ref{sec.gs}.

\subsection{Convex Decomposition for Accurate Collision Detection during Cable Grasping}\label{sec.convex_decomposition}

\begin{figure}[htbp]
	\begin{center}
		\includegraphics[width=7cm]{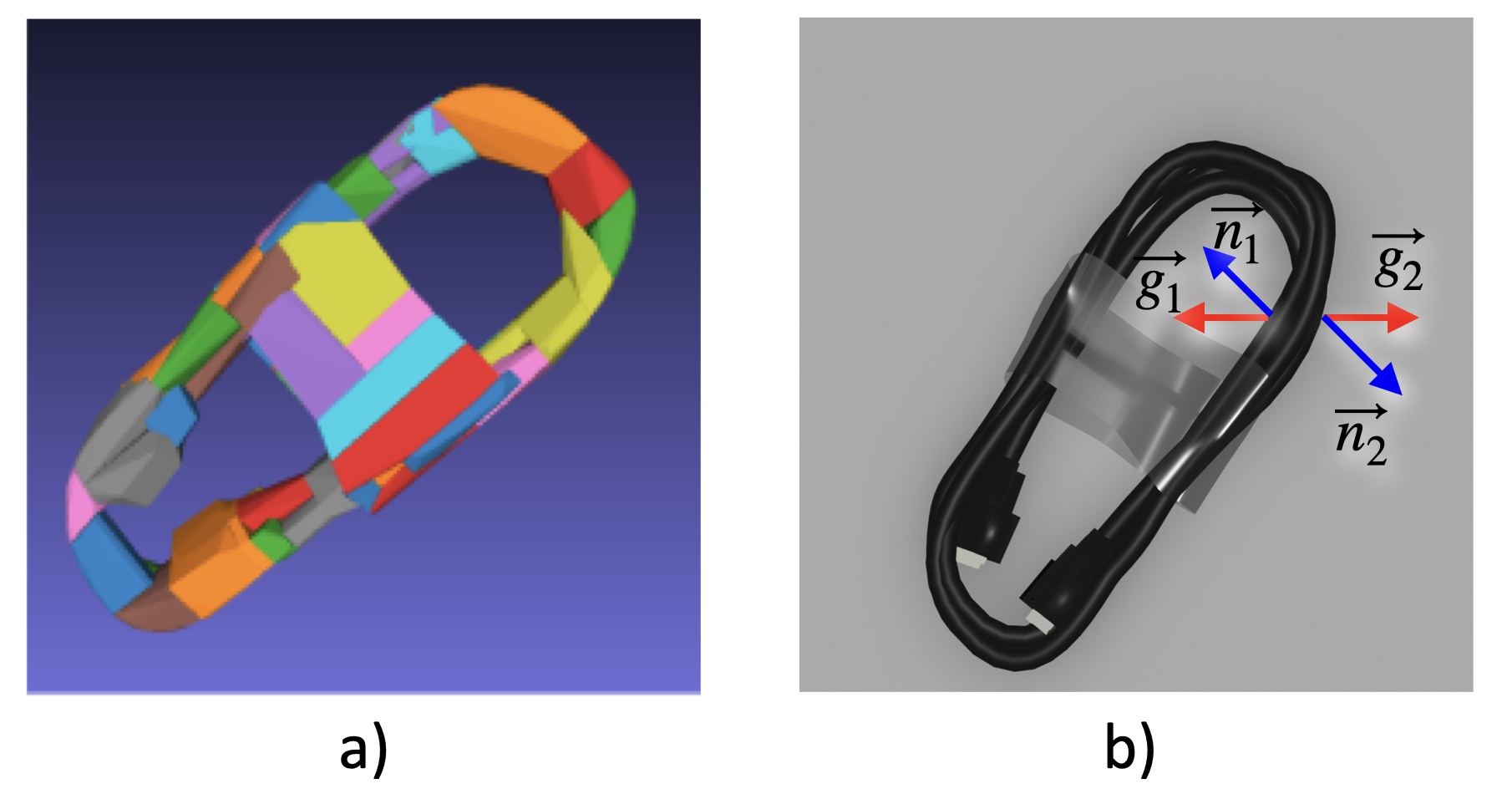}
		\caption{a) To simulate cable grasping with the accurate model of cables instead of convex hull, we decompose the arbitrary polyhedron of the cable with a set of convex shapes. b) Force closure grasp is shown based on normal direction $\Vec{n}_1,\Vec{n}_2$ and grasp directions $\Vec{g}_1,\Vec{g}_2$ of contact points.}
		\label{fig.collision_vis_pybullet}
	\end{center}
\end{figure}

To simulate the accurate collision detection of complicated shapes in PyBullet~\cite{coumans2019}, we typically approximate intricate 3D models $P$ by decomposing them into a group of $M$ smaller convex shapes $[C_{0},C_{1},\ldots,C_{M-1}]$ using volumetric hierarchical convex decomposition (V-HACD) method~\cite{vhacd}, as shown in Fig.~\ref{fig.collision_vis_pybullet} a). This method ensures that the arbitrary polyhedron $P$ is a subset of the union of smaller convex shapes $\bigcup \limits_{i=0}^{M-1} C_{i}$ with the minimal number of shapes $M$.

Convex decomposition method starts by voxelizing the input 3D model, generating approximate convex shapes through hierarchical segmentation based on voxelized data. This segmentation continues recursively until the concavity of each convex shape is sufficiently small. Concavity is determined by evaluating the error between the shapes and their convex hull. Finally, the decomposed mesh is generated from the simplified set of approximate convex shapes by calculating the convex hull of each shape and merging the resulting meshes.

\subsection{Force-Closure Grasping Sampling}\label{sec.force_closure}
Force closure is a concept commonly used to estimate stable grasps and prevent the object from slippery~\cite{nguyen1987constructing}. In this paper, the force-closure grasp sampling algorithm is used to estimate grasp candidates from depth images. The method is described in detailed in Alg.~\ref{alg.grasp_sampling}. 

\begin{algorithm}
	\caption{Grasp Sampling Algorithm}
	\label{alg.grasp_sampling}
	\begin{algorithmic}[1]
		\State Input: Depth image $\boldsymbol I$, Number of sampling $n$, maximal gripper width $w\_max$, friction coefficient $f$
            \State Output: Set of cropped depth images $\bigcup \limits_{j=0}^{n-1} \boldsymbol I_{j}^{\rm c}$ and grasp candidates $\bigcup \limits_{j=0}^{n-1} \textbf{g}_{j}$
		\Repeat
		\State Downsample depth image and crop depth image to speed up the algorithm.
            \State Apply bilateral filter into depth image.
		\State Edge detection according to depth gradient and normal direction estimation according to surrounding edge points.
            \State Sample point pairs $\bigcup \limits_{k=0} (c_{k1},c_{k2})$ from edge points with maximal gripper width $w\_max$. Each point pair contains of two contact points of two-jaw gripper.
            \State Sample grasp directions in set of point pairs with friction coefficient $f$ according to Equation~\ref{equ.force_closure} to obtain valid force-closure grasp candidates, also named antipodal grasp candidates. The grasp width is calculated based on distance between contact points.
		\Until Successfully sample force-closure grasp candidates $\bigcup \limits_{j=0}^{n-1} \textbf{g}_j = \bigcup \limits_{j=0}^{n-1} (x_{j},y_{j},z_{j},\theta_{j},w_{j})$

        \State Crop the original depth image to obtain localized depth images $\bigcup \limits_{j=0}^{n-1} \boldsymbol I_{j}^{c}$, as described in Sec.~\ref{sec.force_closure}.

	\end{algorithmic}  
\end{algorithm}
\setlength{\textfloatsep}{5pt}

\setlength{\textfloatsep}{5pt}

Initially, we detect edge points based on the depth gradients, and estimate the normal directions of these edge points based on the surrounding edge points. Then, we employ a simplified version of force closure method to sample stable antipodal grasp candidates from edge points, as depicted in Fig.~\ref{fig.collision_vis_pybullet} b). The grasp sampling method is represented by the following formula:
\begin{equation}\label{equ.force_closure}
\begin{matrix}
\rm arccos(\textbf{n}_1 \cdot (-\textbf{g}_1))< \rm arctan \textit{(f)}\\
\rm arccos(\textbf{n}_2 \cdot (-\textbf{g}_2))< \rm arctan \textit{(f)}
\end{matrix}
\end{equation}

Here, $\textbf{n}_1$ and $\textbf{n}_2$ represent the normal directions of sampled contact points and $\textbf{g}_1$ and $\textbf{g}_2$ denote the grasping directions. The friction coefficient is denoted by $f$. Given the estimated normal directions and the defined friction coefficient, grasp orientation $\theta$ is calculated based on grasp directions. The central point of contact points denotes grasp central point $(x, y, z)$ and the distance between contact points represents grasp width $w$. Finally, depth images with localized geometric features are cropped with a fixed shape based on position $(x, y)$ of the grasp central point and orientation $\theta$. The horizontal direction of the cropped depth image represents the grasping direction, and the center of the image represents the grasping central point.

\subsection{CG-CNN training}\label{sec.learn_cnn_from_simulation}

EfficientNet~\cite{tan2019efficientnet} achieved good performance and required only few number of parameters by balancing the depth, width and input image size of the convolutional network. Inspired by this, we design the CG-CNN using the EfficientNet framework. As depicted in Fig.~\ref{fig.pipeline_proposed_method}, the network architecture is comprised of inverted residual blocks named MBConv layers~\cite{tan2019efficientnet}, convolutional layers, pooling layers and a fully-connected layer (FCN). CG-CNN takes the cropped depth image of the area around the grasping location as input data and estimates grasp quality. 

We design an adaptable weight cross-entropy loss function as follows:
\begin{equation}\label{equ.loss_function}
L = - \sum_{m=1}^{M} \phi_{m} \times (y \times \log \widehat{y} + (1 - y) \times \log (1-\widehat{y}))
\end{equation}
where, $y$ and $\widehat{y}$ denotes the ground truth and the inferred grasp qualities. The adaptive weight $\phi_m$ is modified according to label-quantity distribution.

\subsection{Grasping strategy}\label{sec.gs}

Several grasp strategies are studied in this paper:

\begin{itemize}
    \item Random Sampling Policy $\pi_{\rm random}$: The executed grasp candidate is randomly selected from grasp candidates based on the grasp sampling method.
    \item CG-CNN Policy $\pi_{\rm CG-CNN}$: The grasp candidate with highest quality is selected based on inferred grasp qualities $Q_{\rm CG-CNN}$, as shown in Equ.~\ref{equ.hybrid_policy}
    \begin{equation}\label{equ.hybrid_policy}
        \textbf{g}_{\rm opt}= \max_{\textbf{g}} \left\{ \bigcup_{i=0}^{N-1} q_{i} + \lambda * \bigcup_{i=0}^{N-1} (1 - r_{\rm height}\times1/N) \right\}
    \end{equation}
    where, $\lambda$ indicates the weight, and $r_{\rm height}$ denotes sort index of the height of grasp central point in all sampled grasps. 
\end{itemize}

\section{Experiment}
\label{sec:experiment}

\subsection{Experimental Setup}
The simulation and physical experiment setup are shown in Fig.~\ref{fig.pipeline_proposed_method}. In simulation, an UR5e robot mounted with a designed two-jaw gripper is used to grasp cables from a bin box filled with cables. In the real experiment, we use the same robotic arm and gripper, and for perception we use a Photoneo PhoXi 3D Scanner M to extract point cloud, depth and gray images. The scanner resolution is $2064 \times 1544$. 

\begin{figure}[htbp]
	\begin{center}
		\includegraphics[width=8.5cm]{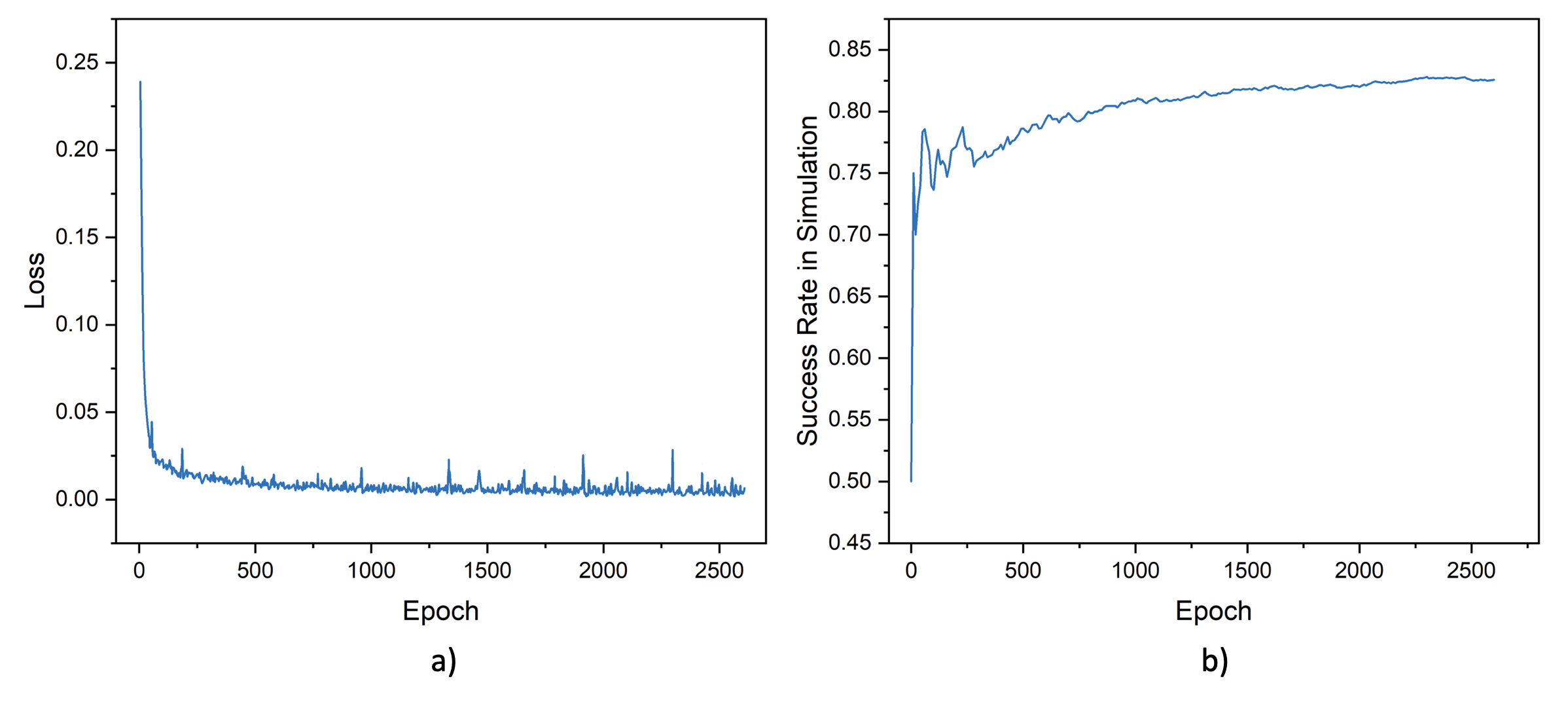}
		\caption{a) Train loss curve. b) Success rate curve of grasping evaluation in simulation.}
		\label{fig.loss_curve_validation}
	\end{center}
\end{figure}
\setlength{\textfloatsep}{5pt}

\subsection{Cable Grasping Dataset}
In simulation, a random number of cables are dropped into a bin from random positions and orientations to generate cluttered scenes. Then, grasp sampling method is used to estimate grasp candidates and the depth images are cropped based on grasp configurations of grasp candidates. UR5e robot is controlled to grasp cables from clutter with accurate collision detection based on convex decomposition. During the grasp simulation process, instances where the grasped object exhibit significant positional deviation are reset. This occurs because the grasped object has not attained a stable pose, and the robot attempts to grasp it prematurely. If the cable is successful grasped, the grasp candidate is labeled with grasp quality - 1; otherwise, it is labeled with grasp quality - 0. Finally, a synthetic cable grasping dataset consisting of over 15,000 data samples is generated with grasp validation results from simulation. 

Compared with Dex-Net, GraspNet-1 billion, and Contact-GraspNet's grasping dataset, where unverified grasp quality is obtained analytically, our dataset's grasping labels are obtained by simulating grasping in simulation, which is more reliable.
\subsection{Grasping Evaluation of CG-CNN in Simulation}

Training CG-CNN is executed by feeding cropped depth images into the network and estimating grasp qualities as the probabilities of grasping successfully. The image size is (224, 224, 3) and the input images are augmented with data augmentation, including horizontal and vertical flipping operations. We use the Adam optimizer~\cite{zhang2018improved} during model training. 
In order to bridge the gap between the physics simulation and the real-world, we adopt the domain randomization method incorporating random initial poses of cables, random friction coefficients, sampled from 0.1 to 0.5, and random salt-and-pepper noise and Gaussian noise.

Grasping evaluation of CG-CNN is executed in proposed simulation environment for every 10 epochs of training. We select the success rate of 100 grasps in simulation as evaluation metric -- dividing the number of successes by the total number of attempted grasping. The loss curve and grasp successful rate from simulation are demonstrated in Fig.~\ref{fig.loss_curve_validation} to show the successful convergence of CG-CNN. These convergences are strong indicators that the network is successful in learning to estimate grasp quality from the input data and effectively generalizing to new examples.

\begin{figure}[htbp]
	\begin{center}
		\includegraphics[width=7.3cm]{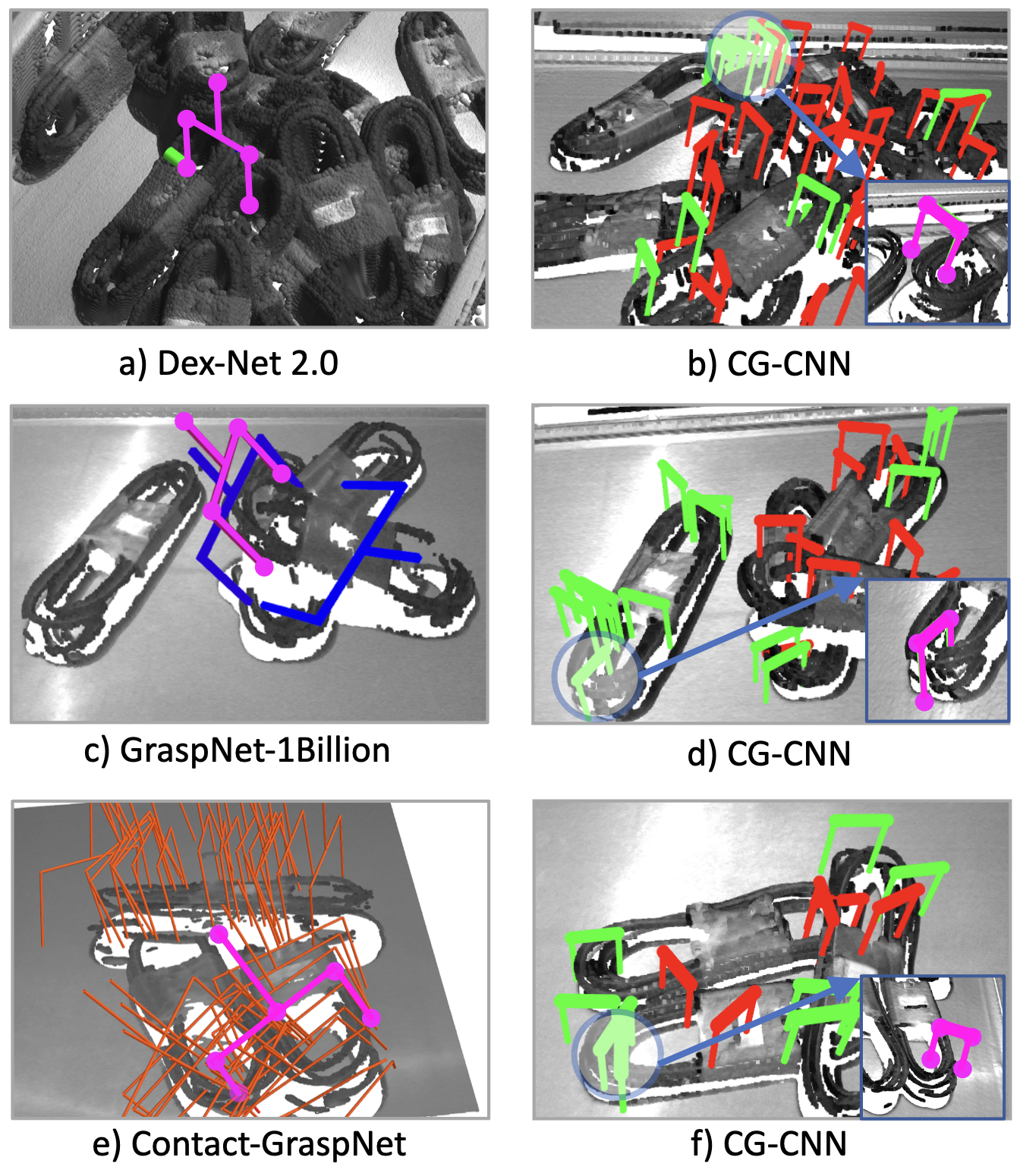}
		\caption{Real-world inference result of Dex-Net 2.0, GraspNet-1Billion, Contact-GraspNet, and our proposed CG-CNN. Optimal grasp is plotted in pink. a) Multi-cable grasping. c) GraspNet-1Billion ignored single cable placed alone and flat on the table, and chose a pose where the object was prone to falling as the optimal grasp candidate. e) Multi-cable grasp. b), d) and f) Optimal grasp in same scenes using CG-CNN.}
		\label{fig.inference_result_collision_graspnet_and_our_cgcnn}
	\end{center}
\end{figure}
\setlength{\textfloatsep}{5pt}

\subsection{Qualitative Study with SOTA Methods}
\label{subsection:qualitative_study_sota}
Qualitatively, we compare the performance of our proposed grasping method with SOTA methods using the same cable grasping scenarios. SOTA methods include Dex-Net 2.0~\cite{mahler2017dex}, graspnet baseline from GraspNet-1Billion~\cite{fang2020graspnet}, and Contact-GraspNet~\cite{sundermeyer2021contact}. Example scenarios with inference results are shown in Fig.~\ref{fig.inference_result_collision_graspnet_and_our_cgcnn}.

CG-CNN is capable of identifying collision-free optimal grasp candidates successfully, as shown in subfigures b), d), and f). This is attributed to CG-CNN's implicit learning of grasp poses and collision conditions in the grasp sampling and evaluation processes from simulation to real world.

Dex-Net 2.0 exhibits the limited capability in cluttered cable grasping scenarios. Collision conditions of grasp candidates are not considered during its data generation. This leads to that multi-cable grasping is selected as optimal grasp by Dex-Net 2.0, as shown in subfigure a). The optimal grasp in same scene in CN-CNN is shown in subfigure b).

The graspnet baseline of GraspNet-1Billion may select dropped grasp sample as optimal grasp, as illustrated in subfigure c). This occurs because GraspNet-1Billion's grasp samples do not undergo grasp verification. It primarily filters grasp samples from single-object environments in the context of complex environments. This limitation hampers its ability to adequately assess grasp quality. Furthermore, we have observed that this method performs poorly in sampling grasps for thin objects due to its exclusion of grasp samples that collide with the tabletop surface. Consequently, the model cannot generate grasp samples for thin objects and loses its capacity to evaluate the quality of grasp samples in collision-prone scenarios. In contrast, CG-CNN validates grasps in a simulation environment across all conceivable conditions, thereby acquiring the ability to evaluate the grasp samples considering environmental collisions. e.g. subfigure d).

Contact-GraspNet frequently fails when attempting to grasp multiple cables, as illustrated in the subfigure e). This occurs because it cannot accurately differentiate whether the cables are situated on multiple objects. To address this issue, CG-CNN marks samples where the number of grasped objects exceeds one during the simulation as failed grasps. Its model predicts the optimal grasp in the same environment, as depicted in the subfigure f).

In conclusion, based on the qualitative experiments, our method is suitable for high-performance cable grasping in cluttered scenes. Through model optimization within simulated environments, we can efficiently and effectively learn the grasp evaluation with implicit collision information. This contributes to the reduction of grasping multiple cables and instances of cable dropping.

\subsection{Quantitative Grasping Experiments}
\subsubsection{Ablation study of CG-CNN and Random Sampling policy}

Quantitatively, we evaluate the CG-CNN and compare it with random sampling policy on successful grasping  rate in simulation and real-world. Collection of grasp success rate is accomplished by dropping different numbers of cables into bin box and performing 100 grasps in the simulation or 50 grasps in the real world. We specially take care the multi-cables grasping. Any cases of grasping multiple cables and dropping after grasping are counted as failures.

\begin{figure}[htbp]
	\begin{center}
		\includegraphics[width=8cm]{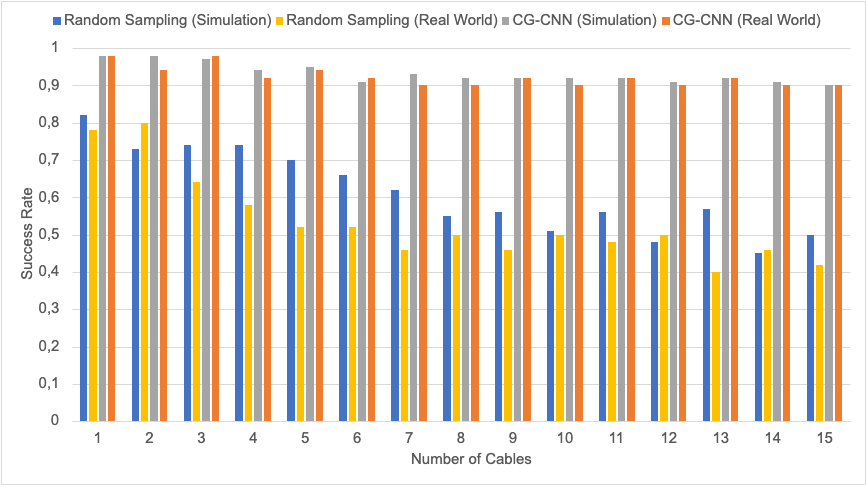}
		\caption{Ablation study results of random sampling and CG-CNN in simulation and real world.}
		\label{fig.exp_result_with_random_sampling_sim_to_real}
	\end{center}
\end{figure}

The successful grasping rate with different numbers of cables are summarized in Fig.~\ref{fig.exp_result_with_random_sampling_sim_to_real}. Our proposed CG-CNN achieve the best performance with an average successful rate of 92.3\% and 93.1\% for cable grasping from a complex stacked environment respectively in real world and simulation. In contrast, the performance of random sampling policy is worse due to factors such as grasping multiple cables, selecting the wrong grasp orientation, and dropping after gripping. Overall, CG-CNN is able to sample and evaluate cable grasp candidates from clutter with high performance both in simulation and real world, and outperformed random sampling policy by 38.8\% in real world.

\subsubsection{Comparison experiments of CG-CNN and SOTA methods}

\begin{figure}[htbp]
	\begin{center}
		\includegraphics[width=8cm]{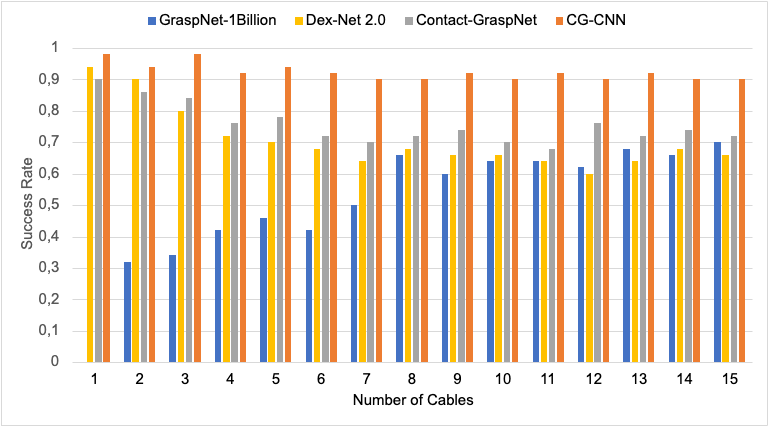}
		\caption{Comparison experiment results of SOTA methods and CG-CNN in real world.}
		\label{fig.exp_result_with_sota_real_world}
	\end{center}
\end{figure}

We conduct quantitative comparison between CG-CNN and the SOTA methods described in~\ref{subsection:qualitative_study_sota}, where results are summarized in Fig.~\ref{fig.exp_result_with_sota_real_world}. Our method exhibits superior performance in cluttered scenarios with varying numbers of cables comparing with the current SOTA method. The proposed approach achieves a successful rate exceeding other methods by at least 16\%. When evaluating grasp sampling using GraspNet-1Billion and Contact-GraspNet, grasp pose is insensitivity to grasp angles. This often leads to post-grasp drops even for optimal grasps. Furthermore, through quantitative experiments, we observe that GraspNet-1Billion is unable to generate grasps for thin objects. The performance degradation of Dex-Net is caused by the lack of implicit collision data in the grasp evaluation process. The experimental results demonstrate that CG-CNN reduce failures of multi-cable grasping and cable dropping, surpassing currently popular algorithms.


\subsubsection{Generalization of CG-CNN}

\begin{figure}[htbp]
	\begin{center}
		\includegraphics[width=3.3cm]{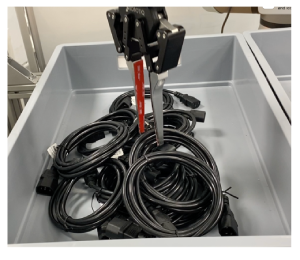}
		\caption{Grasping unseen cable from clutter.}
		\label{fig.grasp_unknown_cable}
	\end{center}
\end{figure}

To assess the generalization performance of our approach, we deploy the trained CG-CNN in grasping unknown cables with varying geometric characteristics, as illustrated in the Fig.~\ref{fig.grasp_unknown_cable}. Through physical experiments, we achieve 88.4\% successful rate in grasping. The results of the experiments demonstrate the generalization capability of our method across cables with diverse shapes.

\section{Conclusions and Future Work}
\label{sec:conculsion}

We present a collision-aware cable grasping network: CG-CNN to learn grasping cables in cluttered environment. A physics simulation scenario is developed to emulate cable grasping. Grasp candidates are sampled with analytical grasp sampling method and the cropped depth images are fed into CG-CNN for training from simulation and inferring grasp quality. The best grasp will be selected by grasping strategy and deployed for implementing grasp experiments.

The feasibility of CG-CNN has been substantiated through simulation and real-world experiments with cluttered scenes featuring various numbers of cables. Qualitatively and quantitatively, comparative experiments have demonstrated our proposed CG-CNN's reduces the cases of grasping multiple cables as well as cable dropping. It outperforms SOTA methods, including Dex-Net 2.0, GraspNet-1Billion and Contact-GraspNet, achieving an average grasp success rate of 92.3\%. Furthermore, our approach exhibits a notable capacity for generalization, achieving an 88.4\% grasp success rate when grasping unknown cables. 

We deployed our methodology in a monitor assembly factory, where the algorithm's performance and speed met industrial standards and commercial requirements, facilitating the real-world implementation of robotic grasping techniques. In future endeavors, we intend to explore cable grasping employing multi-fingered robotic hand.
%

\bibliographystyle{IEEEtran}
\bibliography{main}

\end{document}